\documentclass[journal]{IEEEtran}
\usepackage{cite}
\usepackage{amsmath,amssymb,amsfonts}
\usepackage{algorithmic}
\usepackage{graphicx}
\usepackage{textcomp}
\usepackage{xcolor}
\usepackage{hyperref}
\usepackage{booktabs}
\usepackage{multirow}
\usepackage{float}
\usepackage{amsthm}
\newtheorem{MyDef}{Definition}

\begin{document}
\title{Heterogeneous Edge-Enhanced Graph Attention Network For Multi-Agent Trajectory Prediction}
\author{Xiaoyu Mo,
        Yang Xing,~\IEEEmembership{Member,~IEEE,}
        and~Chen~Lv,~\IEEEmembership{Senior Member,~IEEE,}
\thanks{Xiaoyu Mo and Chen Lv are with the School of Mechanical and Aerospace Engineering, Nanyang Technological University, 639798, Singapore. (e-mail: XIAOYU006@e.ntu.edu.sg, lyuchen@ntu.edu.sg)}
\thanks{Yang Xing is with the Department of Computer Science, University of Oxford, OX1 3QD, UK. (e-mail: yang.xing@cs.ox.ac.uk)}
}

\maketitle

\begin{abstract}
Simultaneous trajectory prediction for multiple heterogeneous traffic participants is essential for safe and efficient operation of connected automated vehicles under complex driving situations in real world. The multi-agent prediction task is challenging, as the motions of traffic participants are affected by many factors, including their individual dynamics, their interactions with surrounding agents, the traffic infrastructures, and the number and modalities of the target agents. To further advance the trajectory prediction techniques, in this work we propose a three-channel framework together with a novel \textbf{H}eterogeneous \textbf{E}dge-enhanced graph \textbf{AT}tention network (HEAT), which is able to deal with the heterogeneity of the target agents and traffic participants involved. Specifically, the agent's dynamics are extracted from their historical states using type-specific encoders. The inter-agent interactions are represented with a directed edge-featured heterogeneous graph, and then interaction features are extracted using the proposed HEAT network. Besides, the map features are shared across all agents by introducing a selective gate-mechanism. And finally, the trajectories of multi-agent are executed simultaneously. Validations using both urban and highway driving dataset show that the proposed model can realize simultaneous trajectory predictions for multiple agents under complex traffic situations, and achieve the state-of-the-art performance with respect to prediction accuracy, demonstrating its feasibility and effectiveness.
\end{abstract}


\IEEEpeerreviewmaketitle

\section{Introduction}
\IEEEPARstart{A}{ccurately} predicting future trajectories of moving objects that share road with autonomous vehicles is an important task in the field of self-driving. With predicted trajectories of surrounding agents, autonomous vehicles can make decisions in advance and avoid possible accidents. This increases the safety, efficiency, and comfort of autonomous driving. However, Trajectory prediction is challenging especially in urban driving scenarios since motion of an agent is affected by many factors: its own dynamics, its interaction with neighboring agents, and the road structure. Researchers in the field of autonomous driving have proposed many works for trajectory prediction and these methods fall into three categories: physics-based, maneuver-based, and interaction-aware methods~\cite{lefevre2014survey}. Physics-based methods~\cite{ammoun2009real} considers the object's individual dynamics to predict its motion ignoring possible maneuvers restricted by the road structure and neighboring agents' impacts. Maneuver-based methods~\cite{althoff2009model} considers maneuver options and predict trajectory conditioned on maneuvers ignoring the impact of surrounding vehicles. Interaction-aware methods~\cite{deo2018convolutional, li2019grip, mo2020recog, li2020social} attract more and more interests recently in that: 1) They naturally treat driving as an interactive activity; 2) They show better performance comparing to pure physics-based and maneuver-based methods; 3) They can be extended to take physics and maneuvers into account. Most existing interaction-aware methods represent motion of all agents in a shared coordinate system, which is sensitive to translation and rotation, and aims at predicting trajectory of a single agent~\cite{deo2018convolutional, mo2020interaction, mo2020recog, zhao2019multi, li2020social}. However, autonomous vehicles should simultaneously predict future states of multiple surrounding agents, e.g., vehicles and pedestrians, to navigate in complex and highly dynamic urban driving scenarios. 

This work focuses on simultaneously predicting future trajectories of multiple heterogeneous agents for both urban and highway driving by jointly considering agents' individual dynamics, their interactions, and the road structure. Agents' past states and a top view image of the interested area is assumed to be available leveraging the vehicle-to-vehicle and vehicle-to-infrastructure communications~\cite{talebpour2016influence}. 

This work proposes a three-channel framework for multi-agent trajectory prediction considering agent's individual dynamics, interactions, and road structure in each channel. For agents' dynamics, this work places each agent in its exclusive coordinate system to eliminate the impacts of coordinate shifting since its recorded states, no matter placed in which coordinate system, can be always converted to its own coordinate system, without affecting other agents. For inter-agent interaction, this work represents interaction as an edge-featured heterogeneous graph and proposes a novel heterogeneous edge-enhanced graph attention network to model the interaction among agents of different types. For the road structure, this work shares a pictorial map across all agents with a gated map selector. 

The main contributions of this work can be summarised as:
\begin{itemize}
    \item A three-channel framework is proposed for multi-agent trajectory prediction. It jointly considers agents' individual dynamics, their interactions, and the road structure for trajectory prediction.
    \item A comprehensive and transformation-insensitive interaction representation is proposed based on edge-featured heterogeneous graph, where the nodes and edges fall into different categories and contains corresponding attributes.
    \item A novel heterogeneous edge-enhanced graph attention network (HEAT) is proposed to model the inter-agent interaction for multi-agent trajectory prediction.
    \item A gate-based map selector is proposed to allow sharing the map across all target agents in a selective manner rather than store a local map for each agent or share the same map across all agents.
\end{itemize}

The remainder of this work is structures as follows: Sec.~\ref{sec: relatedworks} introduces existing works most related to this work. Sec.~\ref{sec: structure} provides an overview of the proposed method. Sec.~\ref{sec: method} elaborates the proposed method and its key components. Sec.~\ref{sec: validation} validates the proposed method on real-world driving dataset collected from both urban and highway scenarios. Sec.~\ref{sec: conclusion} concludes this work and outlines possible future improvements.

\section{Related works}
\label{sec: relatedworks}
This section reviews interaction is representation for interaction-aware trajectory prediction, various graph neural networks (GNNs) proposed for graph-based tasks, and how GNNs are applied to trajectory prediction tasks. This distinction and and advantage of the proposed interaction representation, graph neural network, trajectory prediction frame work are illustrated following each group of related works.
\subsection{Interaction Representation}
Interaction-aware trajectory prediction methods have proposed many ways to represent inter-agent interactions recently. Convolutional social pooling~\cite{deo2018convolutional} designs an occupancy grid, where each cell contain features of the agent falls in it, to model the interaction among agents in the grid. The grid representation is modified in~\cite{mo2020interaction} to observe only the eight agents that mostly affects the target vehicle's behavior. The grid representation is applicable to highway driving, since highway is almost straight and can be easily divided into a grid. But this is not the case for highway driving. To model interaction beyond highway driving, multi-agent tensor fusion~\cite{zhao2019multi} models the interaction by aligning agent's individual features to a top-view image of the driving scene. This still ignores the relationships among the agents. More and more works represent interactions as a graph, where each node represents an agent and the edge represents the inter-agent relationship. Authors of~\cite{diehl2019graph} propose to represent the inter-vehicle interaction as a homogeneous directed graph for highway driving, where each vehicle is connected to its up to eight neighbors. The homogeneous graph ignores the type of agents. GRIP~\cite{li2019grip} also uses a homogeneous graph to model the interaction. ReCoG~\cite{mo2020recog} proposes to represent the interaction as a heterogeneous graph, where a node represents either an agent or a map. An agent is connected to other agents within a neighborhood. ReCoG ignores the edge attributes between nodes. VectorNet~\cite{Gao_2020_CVPR} and TNT~\cite{zhao2020tnt} both use a hierarchical heterogeneous graph to represent the interaction, where each object is represented by a sub-graph, and all the objects are then represented by a fully-connected graph. Edge attribute is ignored in both.
SCALE-Net~\cite{jeon2020scale} proposes to represent the interaction with an edge-featured homogeneous graph, where the edge features contains relatives states between two connected agents. It ignores the heterogeneity of traffic participants. 
Social-WaGDAT~\cite{li2020social} proposes to generate a dynamic pair of history and future graph for each time step. The nodes are assumed to be homogeneous and with a fixed number. EvolveGraph~\cite{li2020evolvegraph} learns an interaction graph which considers the heterogeneity of nodes and edges' types and directions. However, the edge attribute is not considered.

Representing inter-agent interaction as a graph is more natural than using image or grid. However, most existing graph representations place all the agents on the same target-centered coordinate system, which is suitable for single-agent trajectory prediction but can hardly generalize to multi-agent situations because of the effects of coordinate translation and rotation. SCALE-Net~\cite{jeon2020scale} places all agents in their own exclusive coordinates system for generalization and uses edge attributes to preserve spatial relationship among agents. But the graph representation in SCALE-Net is not comprehensive to cover the heterogeneity of agents and their relationships for trajectory prediction. This work proposes to represent the inter-agent interaction in exclusive coordinate systems as a directed heterogeneous edge-featured graph, where different agents are represented by different node and the edge between two agents is assigned with both attribute and type.

\subsection{Graph Neural Networks}
Neural Networks have proven their powerful expression ability on the tasks with well structured data, i.e., image classification~\cite{he2016deep} with grid-like data, and machine translation~\cite{bahdanau2014neural} with chain-like data. However, there are many interesting tasks with data represented in the form of graphs~\cite{mo2019effects}. More and more works are proposed to generalize neural networks to the graph domain. These works are either spectral ~\cite{henaff2015deep, defferrard2016convolutional, kipf2017semi} or non-spectral approaches~\cite{monti2017geometric, hamilton2017inductive, velickovic2018graph}. Spectral methods, i.e., graph convolutional network (GCN)~\cite{kipf2017semi}, depend on Laplacian eigenbasis of the graph, which is hard to calculate for a large graph, while non-spectral methods, i.e., graph attention network (GAT)~\cite{velickovic2018graph}, perform information aggregation only on the local neighborhood avoiding heavy calculation of Laplacian eigenbasis. GAT~\cite{velickovic2018graph} is designed for homogeneous graph. It introduces an attention mechanism to the feature aggregation from neighborhood allowing weighting a neighbor node according to node features. Edge is weighted in GAT, but edge attribute is not covered.Edge en- hanced graph neural network (EGNN)~\cite{gong2019exploiting} addressed continuous multi-dimensional edge feature by using an each dimension to guide an individual attention operation. Convolution with Edge-Node Switching graph neural network (CensNet)~\cite{jiang2020co} utilizes the line graph of the original undirected graph, and designed convolution operations on both graphs to explore edge features. NENN~\cite{yang2020nenn} incorporates node-level and edge-level attentions in a hierarchical manner and learns the node and edge embeddings in the corresponding level. EGAT~\cite{chen2021edge} extends GAT with edge embedding to handle continuous edge features of undirecetd homogeneous graphs. Heterogeneity of nodes and edges in a graph is ignored in the above mentioned works. Heterogeneous graph attention network (HAN)~\cite{wang2019heterogeneous} proposes to handle heterogeneous nodes in a graph with a hierarchical attention mechanism, where the node-level attends over meta-path based neighbors and the semantic-level attends over different meta-paths. Heterogeneous Graph Transformer~\cite{hu2020heterogeneous} proposes node- and edge- type dependent attention mechanism to handle both node and edge heterogeneity in a graph followed by heterogeneous message passing mechanism and target-specific aggregation for feature updating. These GNNs handle heterogeneity in a graph but ignore the edge features. For more information about graph neural networks, please refer to recent reviews~\cite{zhou2020graph, wu2020comprehensive}.

Most existing graph neural networks handles heterogeneity and edge features separately and cannot be directly used to model the interaction represented by a directed edge-featured heterogeneous graph. This work extends GAT~\cite{velickovic2018graph} to handle both heterogeneity and edge features for interaction modeling in multi-agent trajectory prediction.

\subsection{Trajectory Prediction With GNNs} Graph-based interaction representation attracts more and more interests in the field of trajectory prediction and it gives rise to the application of graph neural networks on trajectory prediction. Authors of~\cite{diehl2019graph} test two widely used GNNs (GCN~\cite{kipf2017semi} and GAT~\cite{velickovic2018graph}) and their adaptions on the trajectory prediction task and find that adaptions of GNNs, which discern between the target and surrounding agents, outperforms the GNNs that treat them without distinction. They conceptually prove the effectiveness of graph-based interaction representation but the agents dynamics is note considered in their work. GRIP~\cite{li2019grip} proposes graph a convolutional model, which comprises convolutional and graph operation layers alternatively, to summarize the temporal and spatial features of interactive agents. The convolutional layer is applied to the temporal dimension and the graph operation is applied to spatial relationships. Then an LSTM-based encoder-decoder is used for final prediction. GRIP can predict trajectories of multiple agents but it ignores edge attributes. SCALE-Net~\cite{jeon2020scale} constructs edge attributes with the relative measurements between two agents and employs Edge-enhanced Graph Convolutional Neural Network~\cite{gong2019exploiting} to summarize interactions considering edge attributes. Above mentioned models ignore the heterogeneity of traffic participants. Socail-WaGDAT~\cite{li2020social} designs Wasserstein Graph Double-Attention Network to learn the structure of the interaction graph dynamically and applies kinematic constraints on the predicted trajectory. VectorNet~\cite{Gao_2020_CVPR} applies GNN to a fully-connected hierarchical graph, where a sub-graph contains the feature of an object (either an agent or map component) represented by a sequence of vectors. Heterogeneity is considered in the constructed graph, but the fully-connected graph ignores the spatial structure of interaction and the number of edges increase exponentially with the number of nodes. TNT~\cite{zhao2020tnt} adopts VectorNet as interaction feature extractor and further considers the multi-modality of driving by predict multiple trajectories conditional on selected target points in the map. It shares the drawbacks of VectorNet. ReCoG~\cite{mo2020recog} constructs a heterogeneous graph to represent agent-agent and agent-infrastructure relationships, where the infrastructure (a top-view map) is a node in the graph, and applies GAT and GCN to extract interaction features. Edge attribute and type are ignored in ReCoG. 

Existing trajectory prediction methods are proposed for a specific interaction representations and their can hardly be applied to new representations.

\section{Structure Overview}
\label{sec: structure}
\begin{figure}
    \centering
    \includegraphics[trim={0cm 0cm 0cm 0cm}, clip, width=0.48\textwidth]{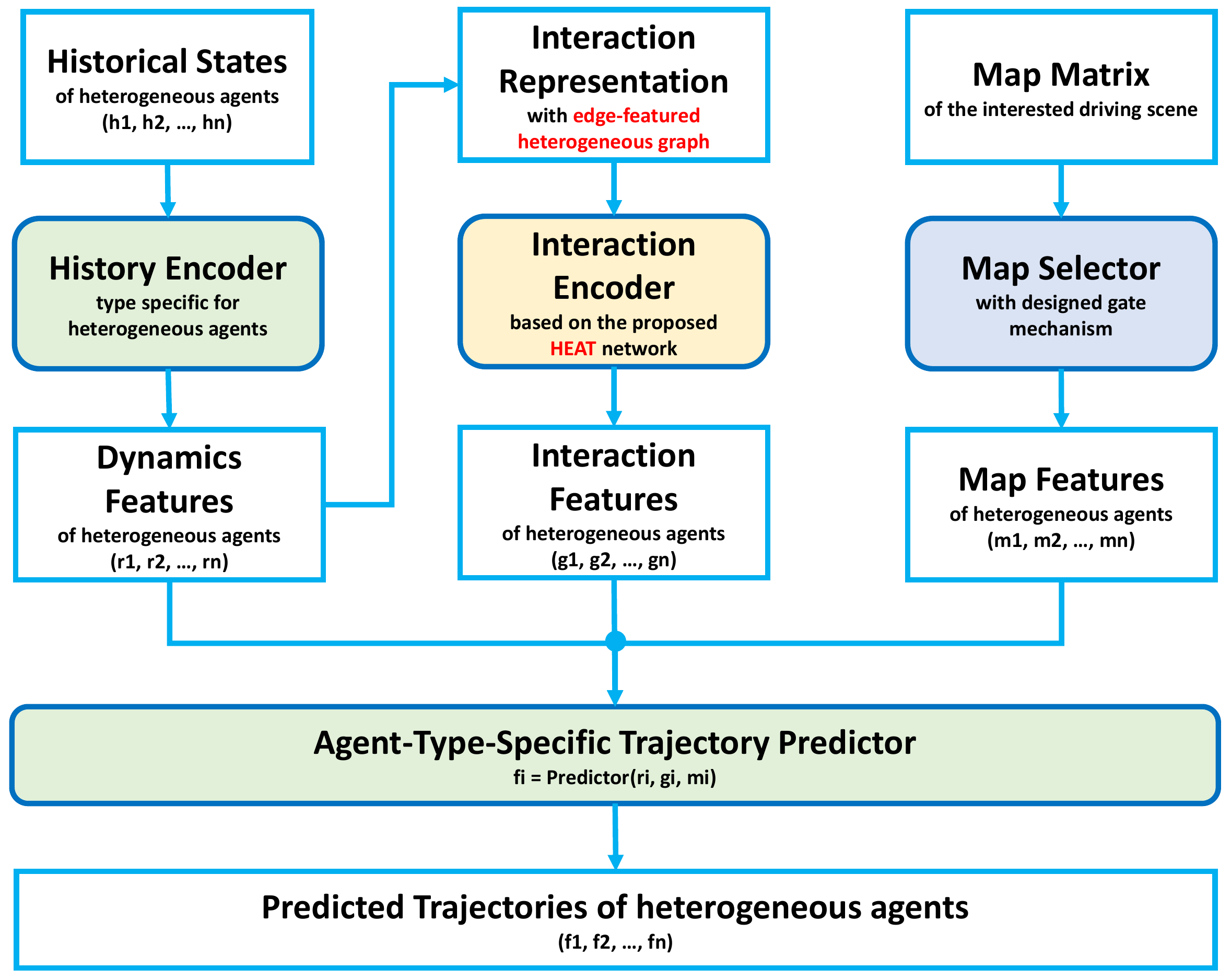}
    \vspace{-2mm}
    \caption{\textbf{Proposed multi-agent trajectory prediction framework.} Historical states of agents are encoded with type-specific history encoders to get their individual dynamics features. Inter-agent interaction is represented by an edge-featured heterogeneous graph with each node contains the dynamics feature of its corresponding agent. Then the proposed HEAT is applied to the interaction graph to extract interaction features for all agents in parallel. Map feature for an agent is processed with a gate-mechanism by considering its dynamics on the map. Then features from these three channels are concatenated and fed to agent-type-specific trajectory predictor to predict future trajectories of all agents. }
    \label{fig: modelstructure}
\end{figure}

This section introduces the high-level structure of the the proposed framework for heterogeneous multi-agent trajectory prediction and its key components, namely, agent-type-specific history encoder, HEAT-based heterogeneous interaction encoder, adaptive map selector, and agent-type-specific trajectory predictor. The proposed framework has three channels for dynamics, interaction, and map features, respectively, then jointly considers these features to predict future trajectories of heterogeneous agents. See Fig.~\ref{fig: modelstructure} for an illustration of this three channel framework.

\textbf{Input and output.}
\begin{figure*}
    \centering
    \includegraphics[trim={0cm 0cm 0cm 0cm}, clip, width=1.0\textwidth]{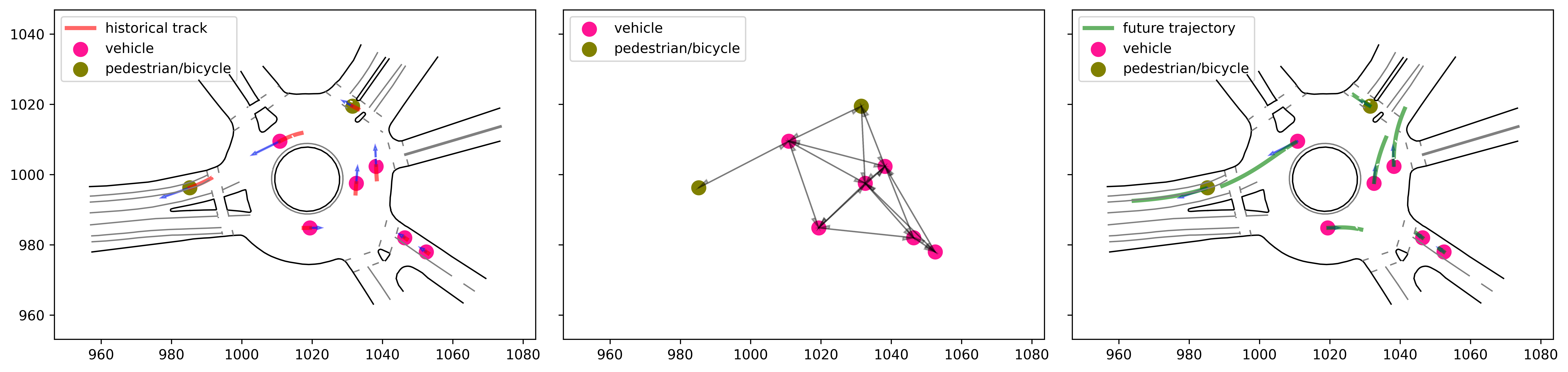}
    \vspace{-8mm}
    \caption{\textbf{Input, graph, and output.} \textit{Left}, the one-second historical tracks of multiple agents of different types navigating in a roundabout scene. \textit{Middle}, The structure of the constructed directed heterogeneous graph with neighboring connections. Self-loop is masked out for clarity. \textit{Right}, the three-second future trajectories of multiple heterogeneous agents in the scene. The pink and olive-green dots show the current positions of vehicle and pedestrian/bicycle agents, respectively. The red solid lines in the left figure are the historical trajectories of the agents over the last one second. The green solid lines in the right figure are the corresponding future in three seconds. This figure is sampled from a roundabout scenario named \textit{DR\_USA\_Roundabout\_FT} in the INTERACTION dataset.}
    \label{fig: inputgraphoutput}
\end{figure*}
The task (see Fig.~\ref{fig: inputgraphoutput} for an illustration) of this work is to simultaneously predict multi-agent trajectories of a group of heterogeneous interactive agents considering their inter-agent interactions and the scene context (shown in the left of Fig.~\ref{fig: inputgraphoutput}). At a time $t$, the input $\mathbf{X}_t$ contains each agent's historical states and the map of the scene (shown in left of Fig.~\ref{fig: inputgraphoutput}).
\begin{equation}
    \mathbf{X}_t= [\mathcal{H}_t, \mathcal{M}],
    \label{eq: input}
\end{equation}
where $\mathcal{H}_t = \{ h^1_t,  h^2_t, \cdots,  h^n_t \}$ contains the historical states of $n$ agents at time $t$, $\mathcal{M}$ is the scene context. Agent $i$'s historical states at time $t$ is represented by $h^i_t = [s^i_{t-T_h+1}, s^i_{t-T_h+2}, \cdots, s^i_t]$, with $T_h$ as the traceback horizon. The state $s^i_t$, for instance, can be the agent $i$'s position and velocity at $t$. The number of observed agents $n$ is variable from case to case. The map $\mathcal{M}$ is to be shared by all the agents. The output contains predicted trajectories of $m\leq n$ heterogeneous agents (shown in right of Fig.~\ref{fig: inputgraphoutput}): 
\begin{equation}
    \mathcal{F}_t = \{ f^1_t,  f^2_t, \cdots,  f^m_t \},
    \label{eq: futall}
\end{equation}
where $ f^i_t =  [(x^{i}_{t+1}, y^{i}_{t+1}),\cdots, (x^i_{t+T_f}, y^i_{t+T_f})]$ is a sequence of the predicted 2D coordinates of agent $i$ over a prediction horizon $T_f$, $\mathcal{F}_t$ is the set of predicted trajectories of $m$ agents. Please note that the number of target agents $m$ is not necessary to be equal to $n$ and can vary from case to case.

\textbf{Agent-type-specific history encoder.} To handle the heterogeneity of traffic participants, this work proposes to share a history encoder over a specific type of traffic participants. For example, if traffic participants fall into three categories, this history encoder (see left of Fig.~\ref{fig: modelstructure}.) will consist of three type-specific encoders. History encoders are applied to individual vehicles' historical states to extract their dynamics features. The dynamics features are also used in the interaction channel as node features.

\textbf{HEAT-based heterogeneous interaction encoder.} This work represents the interaction among heterogeneous traffic participants with a directed edge-featured heterogeneous graph (see middle of Fig.~\ref{fig: inputgraphoutput} for an illustration and Sub.Sec.~\ref{subsec: interrep} for details) and proposes a novel heterogeneous edge-enhanced graph attention network (HEAT) to extract interaction features (see middle of Fig.~\ref{fig: modelstructure}). Nodes in the graph contain dynamics features of corresponding agents out from the history encoder.

\textbf{Adaptive map selector.} This works designs a CNN to extract road feature from a BEV map of the driving scene and selectively share the map feature  over all target agents according to their current positions, velocities, and yaw angles, by introducing a gate-mechanism (see right of Fig.~\ref{fig: modelstructure}).

\textbf{Agent-type-specific trajectory predictor.} Similar to the history encoder, this work shares a trajectory predictor over a specific type of target agents. For example, if target agents fall into two categories, this trajectory predictor will consist of two type-specific predictors. The predictor jointly considers the target vehicles' dynamics feature out from history encoder, their interaction features out from interaction encoder, and their corresponding map features out from map selector, for simultaneous trajectory prediction. See bottom of Fig.~\ref{fig: modelstructure}.

\section{Method}
\label{sec: method}
This section first provides the architecture of the proposed multi-agent trajectory prediction framework (~\ref{subsec: MATP}), then elaborates on the proposed interaction representation (\ref{subsec: interaction}), the proposed heterogeneous edge-enhanced graph attention network: HEAT (\ref{subsec: heat}), and gated map selector (\ref{subsec: map_slct}) for the multi-agent trajectory prediction framework.

\subsection{Heterogeneous Multi-Agent Trajectory Prediction Scheme}
\label{subsec: MATP}
The framework shown in Fig.~\ref{fig: modelstructure} is proposed for multi-agent trajectory prediction, where the agents may have different types, leveraging both historical states of agents and the infrastructure information. To handle the heterogeneity of agents, this work designs specific encoders (\ref{subsubsec: hist_enc}) and decoders (\ref{subsubsec: fut_dec}) for each type of agent. Considered agents are placed in their own exclusive coordinate system and their interactions are represented by a directed edge-featured heterogeneous graph (\ref{subsec: interaction}). A novel heterogeneous edge-enhanced graph attention network is proposed to extract interaction features from the constructed graph (\ref{subsubsec: inter_enc}). To utilize the road structure and share it across all considered agents, this work proposes an adaptive map selector (\ref{subsubsec: map}).
\subsubsection{Agent-Type-Specific History Encoder}
\label{subsubsec: hist_enc}
For an agent of type $\kappa$, its historical states $h^i_t$ is represented by a temporal sequence and can be passed to a type-specific encoder to extract its dynamics feature. RNNs, e.g., Long short-term memory (LSTM) and gated recurrent unit (GRU) are widely used for sequence modeling in machine translation~\cite{cho2014learning, vaswani2017attention} and trajectory prediction~\cite{deo2018convolutional, mo2020recog}. This work adopts GRUs as history encoders (Eq.~\ref{eq: histenc}) because of its effectiveness and simplicity. 
\begin{equation}
    r^i_t = \mathrm{GRU^{\kappa}_{hist}}(h^i_t),
    \label{eq: histenc}
\end{equation}
where $\mathrm{GRU^{\kappa}_{hist}}$ is the historical encoder of agent type $\kappa$ implemented using GRU and $r^i_t$ is the dynamics feature of vehicle $i$ at time $t$. The the output of this module is the dynamics features of all the agents:
\begin{equation}
    R_t = \{ r^1_t,  r^1_t, \cdots,  r^n_t \},
    \label{eq: dynamicsall}
\end{equation}
The dynamics features $R_t$ also serve as the node features in the graph-based interaction representation.

\subsubsection{Heterogeneous Interaction Modeling With HEAT}
\label{subsubsec: inter_enc}
To comprehensively model the inter-agent interaction among heterogeneous agents, this work represents the interaction as a directed edge-featured heterogeneous graph and proposes a novel Heterogeneous Edge-enhanced graph ATtention network (HEAT) to extract interaction features from the graph representation. Details of the interaction representation and the proposed HEAT can be found in Sub.Sec.~\ref{subsec: interaction} and Sub.Sec.~\ref{subsec: heat}, respectively. 

Agents' dynamics features $R_t$ are put into their corresponding node in the graph. Then the proposed HEAT is applied to the graph to model the interaction features for all agent simultaneously.

\begin{equation}
    G_t = \{ g^0_t,  g^1_t, \cdots,  g^n_t \} = \mathrm{HEAT_{enc}}(R_t, E_t),
    \label{eq: heatone}
\end{equation}
where $E_t$ is the edge set containing edge indexes, edge attributes, and edge types, $g^i_t$ is the interaction feature of agent $i$ at time $t$, and $G_t$ contains interaction features of all agents. 

\subsubsection{Map Selection With Gate Mechanism}
\label{subsubsec: map}
Road structure shapes the motion of agents navigating within an urban scene, so that it is necessary to take into consideration the road structure for trajectory prediction. Previous single trajectory prediction methods use a fixed sized local map centered at the target vehicle's current position. But for simultaneous multi-agent trajectory prediction, this map representations has at least two drawbacks: 1) It needs to save multiple maps for multiple agents. 2) The fixed size map can be either two large for a slow agent or two small for a fast agent. To handle the above mentioned drawbacks, this work proposes a adaptive map selection method that allows to share a global map over all the agents according to their current positions, velocities, and yaw angles.
\begin{equation}
    m^i_t = \mathrm{Selector_{map}}\left(\mathcal{M}, (x^i_t, y^i_t, {v_x}^i_t, {v_y}^i_t, \phi^i_t)\right),
    \label{eq: mapslection}
\end{equation}
where $\mathcal{M}$ is the map and $(x^i_t, y^i_t, {v_x}^i_t, {v_y}^i_t, \phi^i_t)$ is the current position, velocity, and yaw angle of agent $i$.

\subsubsection{Agent-Type-Specific Future Decoder}
\label{subsubsec: fut_dec}
For a agent of type $\kappa$, its future trajectory is predicted using a agent-type-specific future decoder by jointly considering its individual dynamics $r^i_t$, its interaction with other agents $g^i_t$, and the map feature regarding to its current states $m^i_t$.
\begin{equation}
    f^i_t = \mathrm{LSTM^{\kappa}_{fut}}([ r^i_t \| g^i_t \| m^i_t ]),
    \label{eq: futdec}
\end{equation}
where $\mathrm{LSTM^{\kappa}_{fut}}$ is the future decoder shared across agents of type $\kappa$, $[ r^i_t \| g^i_t \| m^i_t ]$ is the concatenation of features, and $f^i_t$ is the predicted future trajectory of agent $i$. 

\subsection{Interaction Representation With Directed Edge-featured Heterogeneous Graph.}
\label{subsec: interrep}
\begin{figure}
    \centering
    \includegraphics[trim={0cm 0cm 0cm 0cm}, clip, width=0.48\textwidth]{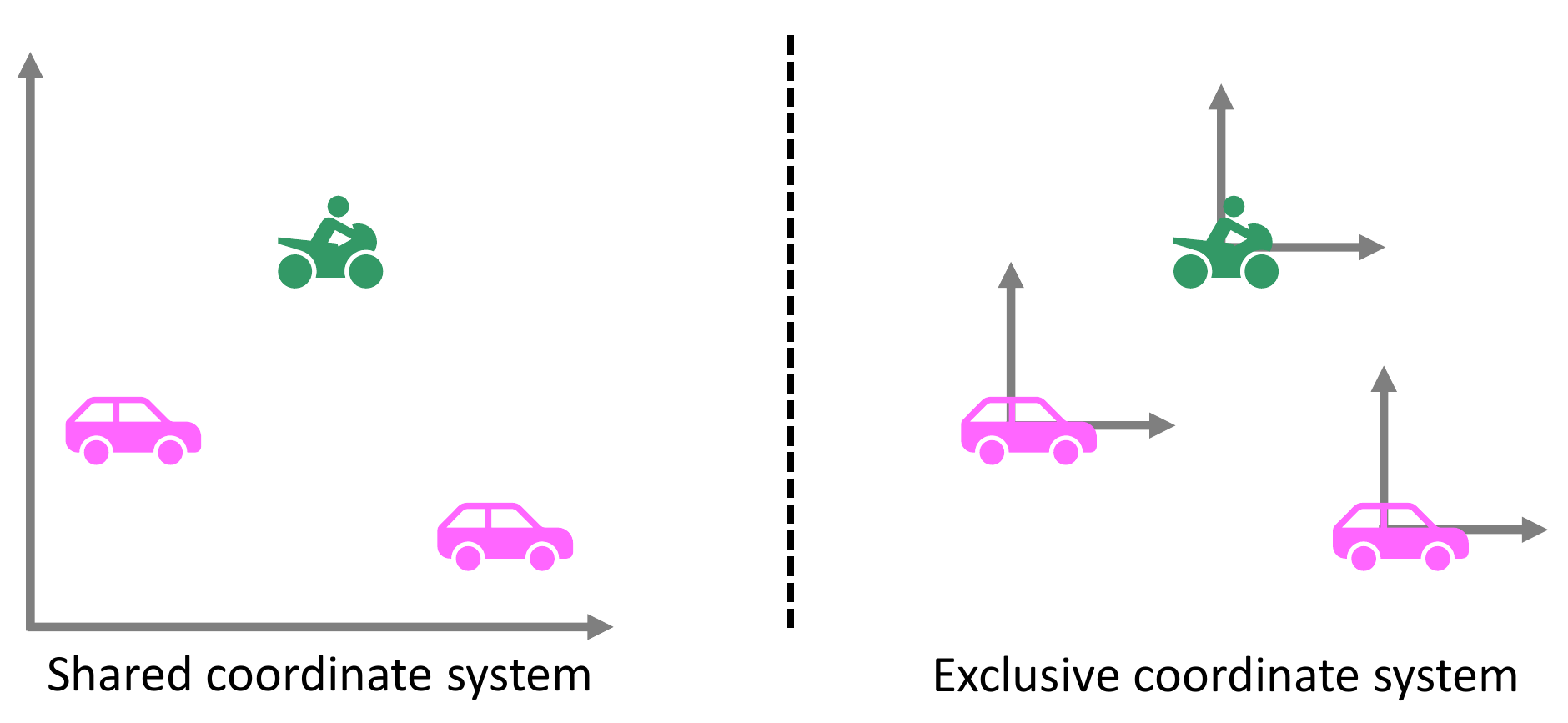}
    \vspace{-2mm}
    \caption{\textbf{Shared and exclusive coordinate systems.} \textit{Left}, Heterogeneous agents in a shared coordinate system. \textit{Right}, Heterogeneous agents with exclusive coordinate systems, with the origin fixed at their current positions and the horizontal axes pointing to their moving directions.}
    \label{fig: coordinatesystem}
\end{figure}

This work places all agents in their own exclusive coordinate systems and represent their interaction as a directed edge-featured heterogeneous graph. 
\label{subsec: interaction}
\subsubsection{Exclusive Coordinate System} Most existing interaction-aware trajectory prediction methods use either a shared coordinate system for all agents or an exclusive coordinate system for each to represent trajectories. A shared coordinate system preserves the spatial relationship among agents, however, it is sensitive to translation and rotation. The input to the model becomes totally different when a new shared coordinate system is applied. But in the case of exclusive coordinate system, agents' 
states are represented locally and independent of other agents. The localized exclusive coordinate system standardises the states of agents but omits spatial relationships among agents, which should be reserved with edge features to take advantage of the exclusive coordinate system.

\subsubsection{Graph Represented Interaction}
This work proposes to represent inter-agent interaction as a directed edge-featured heterogeneous graph for multi-agent trajectory prediction. Each node represents a traffic participant with a specific type and contains its feature extracted from a sequence recorded in its exclusive coordinate system. An edge from node $j$ to node $i$ means that node $i$'s behavior is influenced by node $j$ and the edge attribute is relative measurements of node $j$ to node $i$, e.g., position, velocity, and yaw angle.
The edge type is a concatenation of the type of node $j$ and node $i$. 

\begin{MyDef}[Directed Edge-Featured Heterogeneous Graph]
A directed graph can be represented by $\mathbb{G} = (V, E)$, where $V=\{v_1, \cdots, v_n\}$ is the set of $n$ nodes, and $E \subset V \times V$ is the set of directed edges. Each node contains its node feature and belongs to a specific type. Each directed edge contains an edge attribute and falls into a specific edge type. 
\end{MyDef}

Different to previous works, which represent interaction as a homogeneous graph~\cite{diehl2019graph}, edge-featured homogeneous graph~\cite{jeon2020scale}, or heterogeneous graph without edge attributes~\cite{li2020evolvegraph}, the proposed representation is more comprehensive. It covers the heterogeneity of traffic participants with heterogeneous nodes; preserves their individuality with exclusive coordinate systems; considers the difference of the mutual influence between two agents with directed edges; preserves the spatial relationship of all the agents using edge attributes. 

\textbf{Transformation-insensitive interaction graph.} To model the interaction among agent of different types, this work constructs a directed heterogeneous graph to represent these inter-agent relationships. An edge $e_{ij}$ pointing from agent $j$ to agent $i$ is constructed if agent $j$ is within a predefined neighborhood of agent $i$. Each valid edge $e_{ij}$ is assigned with edge attribute and edge type. Then the edge set is:
\begin{equation}
    E = \{e_{ij}\}_{(j\in \mathcal{N}_i)}, \quad i = 1, \cdots, n,
    \label{eq: edgeset}
\end{equation}
where $i$ and $j$ is the indexes of agents and $\mathcal{N}_i$ is the neighborhood of agent $i$. Self-loop $e_{ii}$ is included in the edge set. An example of the constructed graph is shown in middle of Fig.~\ref{fig: inputgraphoutput}

\subsection{HEAT Layer}
\label{subsec: heat}
The above mentioned interaction representation should be treated with a graph neural network which can handle the heterogeneity of nodes, directed edges, and continuous edge attributes. But, as shown in related works, existing GNNs do not handle this in one fell swoop. This work designs a heterogeneous edge-featured graph attention network (HEAT), an extension of GAT, for the proposed comprehensive interaction representation. HEAT can be constructed by stacking HEAT layers. A HEAT layer updates node features by aggregating information from neighborhoods. It first transforms node and edge features accordingly then aggregates node features via edge-enhanced masked attention (or optional multi-head attention) mechanism.

\subsubsection{Input and Output}
The input to the HEAT layer contains a set of node features: $$\mathbf{h}=\{ \vec{h}_i | i = 1, \dots, n\},$$ where $\vec{h}_i \in \mathbb{R}^{F_h}$ is the feature vector of node $i$; 
A set of edge attributes: $$\mathbf{e}^{attr}=\{ e^{attr}_{ij} | i = 1, \dots, n, j = 1, \dots, N\},$$ where $e^{attr}_{ij}\in \mathbb{R}^{F^{attr}_e}$ is the attribute of the edge pointing from node $j$ to node $i$; In addition, a set of edge types: $$\mathbf{e}^{type}=\{e^{type}_{ij} | i = 1, \dots, n, j = 1, \dots, n\},$$ where $ e^{type}_{ij}\in \mathbb{R}^{F^{type}_e}$ is the type of the edge pointing from node $j$ to node $i$.

It produces a new set of node features: 
$$\mathbf{h}'=\{ \vec{h}'_i | i = 1, \dots, n\}, \vec{h}'_i \in \mathbb{R}^{F'_h}.$$

\subsubsection{Node-Type-Specific Transformation}
Different kinds of nodes in a heterogeneous graph have different feature spaces and should be projected to a shared feature space. This work adopts the node-type-specific transformation matrix $\mathbf{M}_{\kappa i}$ introduced in~\cite{wang2019heterogeneous} to handle arbitrary kinds of nodes. 
$$ \vec{h}_{\kappa i} = \mathbf{M}_{\kappa i} \cdot \vec{h}_i,$$ 
where $\vec{h}_{\kappa i}$ is the projected feature of node $i$ according to its type $\kappa$. 
\subsubsection{Edge Attribute And Type Transformation}
Existing works either consider edge attributes or edge types as edge features, while this is not the case for multi-agent trajectory prediction. We argue that, for trajectory prediction, edge feature and type are two different things in that the edge feature is usually some measurements in a continuous space, such as the distance between two nodes, while edge type is always a discrete indicator. This work jointly consider the edge features and types by introducing the edge attribute transformation:
$$ \mathbf{e}^{attr}_{\phi} = \mathbf{M}_{\phi} \cdot \mathbf{e}^{attr}, $$
where $\mathbf{e}^{attr}_{\phi}$ is the projected edge attributes, and the edge type transformation:
$$ \mathbf{e}^{type}_{\chi} = \mathbf{M}_{\chi} \cdot \mathbf{e}^{type}, $$
where $\mathbf{e}^{type}_{\chi}$ is the projected edge types.

\subsubsection{Edge-Enhanced Masked Attention}
For an edge pointing from node $j$ to node $i$, its edge feature:
$ e_{ij} = [e^{attr}_{\phi ij} \| e^{type}_{\chi ij}]$, is a concatenation of its transformed edge attribute and type. For node $i$, a concatenated feature vector $e^+_{ij}=[e_{ij} \| \vec{h}_{\kappa j}]$ represents the feature of node $j$ from node $i$'s point of view considering the edge attribute and type. $e^+_{ij}$ is then sent to a shared attention mechanism~\cite{bahdanau2014neural}, which is a single-layer feed-forward neural network, $\vec{\mathbf{a}}$, applying LeakyReLU non-linearity and softmax normalization. The attention coefficient $\alpha_{ij}$ indicates the importance of the node $j$ to node $i$ jointly considering node and edge features. GAT layer~\cite{velickovic2018graph} performs masked attention, which attends over the neighborhood of node $i$ only, to utilize the structural information of the graph while casting away the edges' feature and type. This work performs an edge-enhanced masked attention in Eq.~\ref{eq: maskattn} to fully consider the graph attributes.
\begin{equation}
    \alpha_{ij} = \frac{{\rm exp}\left( {\rm LeakeyReLU} \left(\vec{\mathbf{a}}^T [\vec{h}_{\kappa i} \| e^+_{ij}]\right)\right) }{\sum_{k\in \mathcal{N}_i} {\rm exp}\left( {\rm LeakeyReLU} \left(\vec{\mathbf{a}}^T [ \vec{h}_{\kappa i} \| e^+_{ik}]\right)\right) },
    \label{eq: maskattn}
\end{equation}
where $\mathcal{N}_i$ is the neighborhood of node $i$ in the graph. The attention coefficients are then used to update feature of node $i$ with a linear combination over its neighborhood.

\subsubsection{Node Feature Aggregation}
The feature of node $i$ is updated by calculating a weighted sum of edge-integrated node features over its neighborhood, followed by a sigmoid function. 
\begin{equation}
    \vec{h}'_i=\sigma \left(\sum_{j \in \mathcal{N}_i} \alpha_{ij} \mathbf{W}_h [e^{attr}_{\phi ij} \| \vec{h}_{\kappa i}] \right).
\end{equation}
Edge types are not included in the edge-integrated feature since it is discrete and already considered in the previous attention mechanism.

\subsubsection{Multi-Head Attention}
Similar to GAT~\cite{velickovic2018graph}, HEAT allows running several independent attention mechanisms to stabilize the self-attention mechanism. Then the features out from each attention head is concatenated as the updated feature. The so called multi-head attention mechanism is shown in Eq.~\ref{eq: multiheadattn}:
\begin{equation}
    \vec{h}'_i= \mathop{\parallel}\limits^{K}_{k=1} \sigma \left(\sum_{j \in \mathcal{N}_i} \alpha^k_{ij} \mathbf{W}^k_h [e^{attr}_{\phi ij} \| \vec{h}_{\kappa i}] \right),
    \label{eq: multiheadattn}
\end{equation}
where $K$ is the number of attention heads and $\parallel$ represents concatenation.

\begin{table} 
\caption{Notations of HEAT layer}
\centering
\begin{tabular}{l l}
\toprule
$\vec{h}_i$ & Feature vector of node $i$ \\
\hline
$\mathbf{h}$ & The set of node features in a graph \\ 
\hline
$\vec{h}_{\kappa i}$ & Projected feature of node $i$ of type type $\kappa$ \\
\hline
$e^{attr}_{ij}$ & attribute of edge from $j$ to $i$ \\ 
\hline
$\mathbf{e}^{attr}$ & The set of edge attributes in a graph \\ 
\hline
$\mathbf{e}^{attr}_{\phi}$ & projected attribute of edge from $j$ to $i$ \\ 
\hline
$e^{type}_{ij}$ & type of edge from $j$ to $i$ \\ 
\hline
$\mathbf{e}^{type}$ & The set of edge types in a graph \\ 
\hline
$\mathbf{e}^{type}_{\chi}$ & projected type of edge from $j$ to $i$ \\ 
\hline
$e_{ij}$ & concatenation of projected attribute and type \\
\hline
$e^+_{ij}$ & concatenation of $e_{ij}$ and $\vec{h}_{\kappa j}$ \\
\hline
$\alpha_{ij}$ & node $j$'s attention coefficient for node $i$ \\
\hline
$\mathcal{N}_i$ & Neighborhood of node $i$\\
\hline
$\|$ & concatenation \\
\hline
$\vec{\mathbf{a}}^T$ & Attention mechanism \\
\hline
$\sigma$ & Sigmoid function \\
\hline
$\mathbf{h}'$ & The set of updated node features in a graph \\
\bottomrule
\end{tabular}
\end{table}

\subsection{Gated Map Selection}
\label{subsec: map_slct}
The road structure highly affects the motions of traffic participants so that trajectory prediction cannot ignore this information. Single trajectory prediction methods, such as ReCoG~\cite{mo2020recog}, uses a fixed-size local map centered at the target vehicle's current position. A fixed-size local map ignores dynamics of agents. A small map is enough to predict a slow agent, while a larger map is asked for a fast-moving agent.
Multi-agent prediction methods, such as MATF~\cite{zhao2019multi}, share the same map feature across all target vehicles ignoring the fact that different target agents are affected by different parts of the map. To enable selective map sharing, this work proposes to apply the gate mechanism on the CNN-extracted map feature for map selection. The gate mechanism is widely used in sequence modeling~\cite{hochreiter1997long, chung2014empirical, gers2002learning}. LSTMs~\cite{gers2002learning} usually have three gates: input gate, forget gate, and output gate to manage the information flows along the sequence data. The input gate is designed to select what information of the current step to be added to the memory of the network. This work designs the selection gate $z^i_t$ to select map feature for an agent $i$ according to its current state in the map.
\begin{equation}
    z^i_t = \sigma(W_z[\vec{\mathcal{M}} \| s^i_t] + b_z),
    \label{eq: mapgate}
\end{equation}
where $\vec{\mathcal{M}}$ is the map $\mathcal{M}$'s feature vector extracted using a CNN, $s^i_t$ is agent $i$'s current state in the map's coordinates system, $\sigma$ is a sigmoid function, and $z^i_t$ is a vector with each element contains a number between 0 and 1. Then the map feature of agent $i$ at $t$ is selected with this gate. 
\begin{equation}
    m^i_t = z^i_t \circ \vec{\mathcal{M}},
    \label{eq: gatedmapslector}
\end{equation}
where $\circ$ is element-wise production and $m^i_t$ is the selected map feature.

\section{Real-World Dataset Validation}
\label{sec: validation}
This section first compares the proposed trajectory prediction method with state-of-the-art models on the recently published INTERACTION dataset~\cite{interactiondataset} for urban driving, then on the NGSIM US-101~\cite{ushighway101} dataset for highway driving.
\subsection{Validation On Heterogeneous Dataset}
The proposed heterogeneous multi-agent trajectory prediction method is trained and validated on the INTERACTION~\cite{interactiondataset}. The full-name of the dataset is INTERnational, Adversarial and Cooperative moTION Dataset. It contains naturalistic trajectories of different traffic participants, i.e., vehicles and pedestrians, in highly interactive urban scenarios from world-wide. The recorded scenarios fall into three categories: roundabout, intersection, and merging. 
\subsubsection{Heterogeneous Dataset} INTERACTION dataset provides states of agents at each timestamp in along with a high definition (HD) map. The state of a vehicle at a timestamp includes its position, velocity, yaw angle, and shape, while the state of a pedestrian/bicycle includes only its position, velocity, and yaw angle. Since this work aims at simultaneously prediction trajectories of multiple heterogeneous agent and proposes to represent inter-agent interaction as a heterogeneous directed-edge-featured graph, the raw dataset is processed accordingly. A processed data are stored with:
\begin{itemize}
    \item \textit{Historical states}: The historical states within a traceback horizon of all agents. The historical states of agent $i$ (position, velocity, and orientation) is stored in its own exclusive coordinate system with the origin fixed at its current position and the horizontal axis pointing to its current direction. See Fig.~\ref{fig: coordinatesystem} for the illustration of exclusive coordinate system.
    \item Edge indexes: The graph connectivity represented a set of directed edges. A directed edge from node $j$ to node $i$ means that agent $j$ is within the neighborhood of agent $i$ and affects the behavior of agent $i$.
    \item Edge attributes: The attributes of all edges. The attribute of an edge from agent $j$ to agent $i$ contains agent $j$'s relative states to that of agent $i$.
    \item Edge types: The types of all edges. The type of an edge from agent $j$ to agent $i$ contains a concatenation of the types of agent $j$ and agent $i$.
    \item Target masks: The mask of the agents to be predicted. If agent $i$'s future trajectory is to be predicted, its mask is set to $1$, else it's set to $0$.
    \item Vehicle masks: The mask of of the vehicles in a scene with $1$ represents a vehicle and $0$ represents a non-vehicle.
    \item Pedestrian masks: The mask of the pedestrians in a scene with $1$ represents a pedestrian and $0$ represents a non-pedestrian.
    \item Target vehicle masks: The mask of the vehicles to be predicted with $1$ means that the vehicle's future trajectories is to be predicted.
    \item Scene map: The map of the scene represented by a top-view image. Since this work proposes a learned map selector to share the map across all agents, the map can be stored outside a piece of data for just once. That saves a great deal of disk space. A corresponding map is saved for each of the eleven scenarios in the INTERACTION dataset.
    \item Vehicle-to-map attributes: The states of all agent relative to the map's center at current time $t$.
    \item Ground truth future trajectories: The recorded future trajectories of all target agents over the prediction horizon. 
\end{itemize}
The processed dataset is split to train and validation set following the split suggested by the authors of the INTERACTION dataset. The processed dataset contains $425,192$ data pieces for training and $104,627$ for validation.

\subsubsection{Metrics} This work evaluates prediction performance in terms of Average Displacement Error (ADE) and Final Displacement Error (FDE) in meter. ADE and FDE are widely used in previous works~\cite{zhao2020tnt, mo2020recog} and they are calculated as below:
\begin{equation}
    ADE =\frac{1}{T_f}\sum^{T_f}_{\tau=1} \sqrt{(\hat{x}_{\tau} - x_{\tau})^{2} + (\hat{y}_{\tau} - y_{\tau})^{2}},
    \label{eq: ADE}
\end{equation}
\begin{equation}
    FDE = \sqrt{(\hat{x}_{T_f} - x_{T_f})^{2} + (\hat{y}_{T_f} - y_{T_f})^{2}},
    \label{eq: FDE}
\end{equation}
where $(\hat{x}_{\tau}, \hat{y}_{\tau})$ is the predicted position in XY-coordinates at time $\tau$, $(x_{\tau}, y_{\tau})$ is the ground truth of position at time ${\tau}$, and $T_f$ is the prediction horizon.

\subsubsection{Comparison With State-of-the-art methods}
The proposed method is compared with the following methods on the INTERACTION dataset. 
\begin{itemize}
    \item DESIRE: DESIRE predicts multi-modal trajectories by jointly considering motion history, static scene context and inter-agent interactions. It first generates diverse hypothetical future prediction samples using a conditional variational auto-encoder, then ranks and refines the samples in a inverse optimal control framework with regression~\cite{lee2017desire}.
    \item MultiPath: MultiPath handles driving uncertainty by hierarchically model intent and control uncertainties. It first produces a set of $K$ anchor trajectories as intents, then predicts future trajectories conditioned on anchors, where the uncertainty is modeled as a Gaussian distribution given an intent~\cite{chai2019multipath}.
    \item TNT: TNT utilizes VectorNet~\cite{Gao_2020_CVPR} to encode the target agent's interaction with surrounding agents and the environment. It first predicts an agent's target states within a prediction horizon, then generates trajectories for each target. Finally a set of predictions is selected according to estimated likelihoods for multi-modality~\cite{zhao2020tnt}.
    \item ReCoG: ReCoG represents vehicle-vehicle and vehicle-infrastructure interactions as a heterogeneous graph and applies state-of-the-art GNNs for interaction encoding. It predicts a single trajectory for a single target vehicle~\cite{mo2020recog}.
\end{itemize}

\begin{table}
\centering
\caption{\textbf{Comparison with state-of-the-art methods on INTERACTION dataset}}
\centering
\begin{tabular}{|c|c c c|} 
\hline
\textbf{Methods} & MM & ADE@3sec (m) & FDE@3sec (m) \\
\hline
DESIRE~\cite{lee2017desire}  & \checkmark & 0.32 ($\mathbf{min}_6$) & 0.88 ($\mathbf{min}_6$) \\
MultiPath~\cite{chai2019multipath}  & \checkmark
 & 0.30 ($\mathbf{min}_6$)  & 0.99 ($\mathbf{min}_6$) \\
TNT~\cite{zhao2020tnt}  & \checkmark & 0.21 ($\mathbf{min}_6$) & 0.67 ($\mathbf{min}_6$) \\
ReCoG~\cite{mo2020recog}  &  & \textbf{0.19} & \textbf{0.65} \\
\hline
HEAT-I-R (Ours) &  & \textbf{0.19} & 0.66 \\
\hline
\end{tabular}
\vspace{0.2cm}
\label{tab: itothersresults}
\end{table}
Tab.~\ref{tab: itothersresults} compares the proposed three channel model HEAT-I-R with existing methods. It shows that: 1) The proposed HEAT-I-R outperforms DESIRE~\cite{lee2017desire} and MultiPath~\cite{chai2019multipath}, even though these two methods predict multi-modal (MM) trajectories for a single agent and the ADE and FDE are reported with the minimum~\cite{zhao2020tnt}. 2) HEAT-I-R matches the performance of TNT~\cite{zhao2020tnt} and ReCoG~\cite{mo2020interaction}. Please note that TNT~\cite{zhao2020tnt} predicts six-modal trajectories for a single agent and reports the minimum ADE and FDE over all predictions and ReCoG~\cite{mo2020recog}, winner solution of the INTERPRET Challenge (NeurIPS 2020)~\cite{interpretchallenge}, predicts a single trajectory for a single target. Comparing to TNT~\cite{zhao2020tnt} and ReCoG~\cite{mo2020recog}, the proposed HEAT-I-R is able to predict trajectories of multiple agents (MA) simultaneously.

\subsubsection{Ablative Study}
\begin{table}
\centering
\caption{\textbf{Ablative comparison on the INTERACTION dataset's DR\_USA\_Roundabout\_FT scenario}} 
\centering
\begin{tabular}{|c|c c|} 
\hline
\textbf{Methods } & ADE@8sec (m) & FDE@8sec (m) \\
\hline
R & 3.99  & 11.64 \\ 
\hline
GAT & 3.98  & 11.59 \\ 
\hline
GAT-R & 3.5  & 10.62 \\
\hline
HEAT & 3.10  & 8.83 \\
\hline
HEAT-R & 3.09  & 8.84 \\
\hline
HEAT-I-R & \textbf{2.97}  & \textbf{8.56} \\
\hline
\end{tabular}
\vspace{0.2cm}
\label{tab: itablation}
\end{table}

\begin{figure}
    \centering
    \includegraphics[trim={0cm 0cm 0cm 0cm}, clip, width=0.48\textwidth]{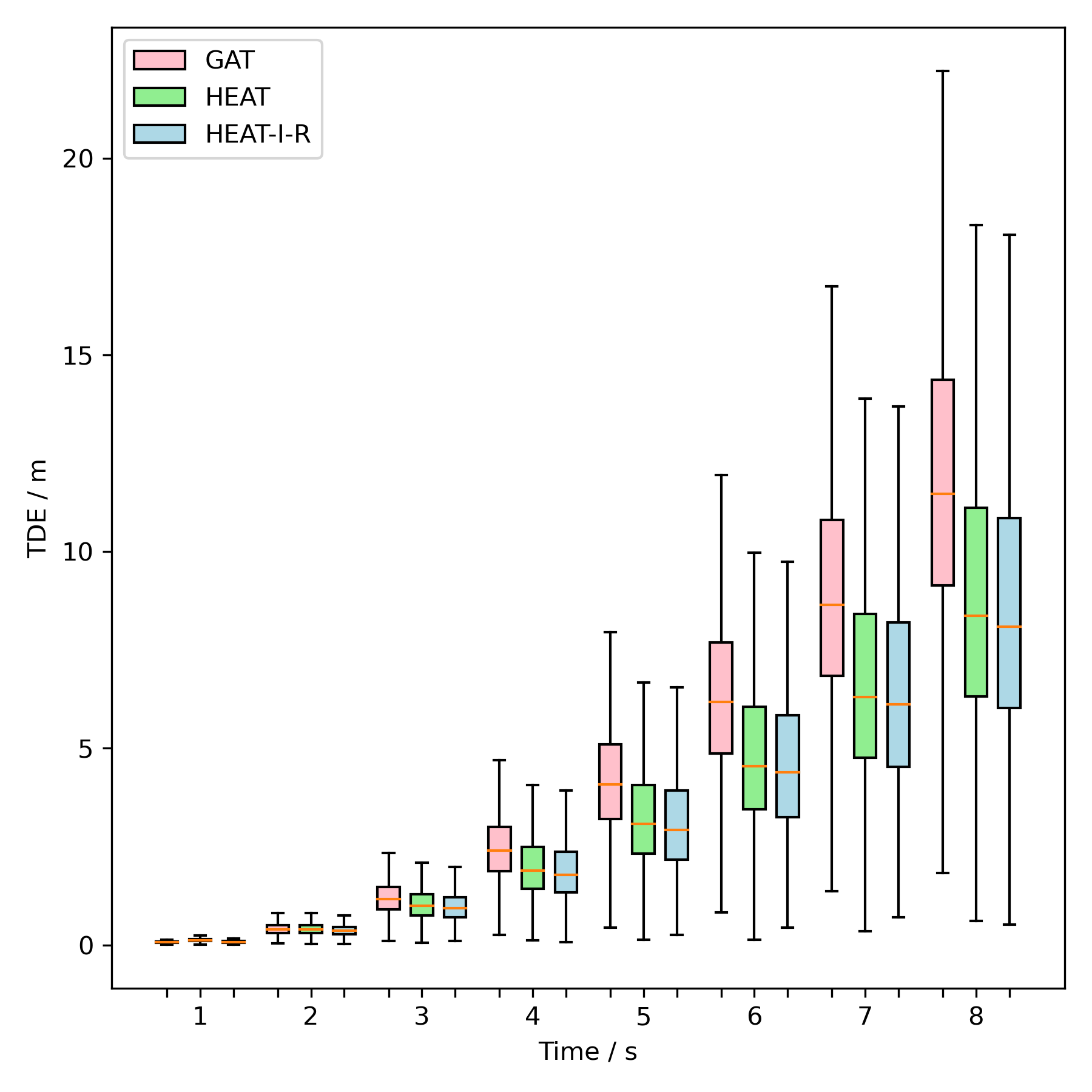}
    \vspace{-2mm}
    \caption{\textbf{Box plots of the TDE of ablative models.} GAT, HEAT, and HEAT-I-R are selected for clarity of the box plots.}
    \label{fig: boxplotFT}
\end{figure}
This work conducts ablative study on the INTERACTION dataset's DR\_USA\_Roundabout\_FT scenario for a longer prediction horizon (8 seconds). The following settings are trained and validated on the same dataset.
\begin{itemize}
    \item R: One-channel model considering the target vehicle's RNN-encoded  dynamics feature only for future trajectory prediction.
    \item GAT: One-channel model predicting the future trajectory of the target vehicle considering its graph-modeled interaction feature extracted using GAT.
    \item GAT-R: Two-channel model jointly considering the GAT-extracted interaction and RNN-encoded dynamics features for trajectory prediction.
    \item HEAT: One-channel model predicting the future trajectory of the target vehicle considering its graph-modeled interaction feature extracted using HEAT.
    \item HEAT-R: Two-channel model jointly considering the HEAT-extracted interaction and RNN-encoded dynamics features for trajectory prediction.
    \item HEAT-I-R: The proposed three-channel frame work, which combines the target vehicle's individual dynamics feature, its interaction feature, and the selected map feature for trajectory prediction.
\end{itemize}

Tab.~\ref{tab: itablation} shows the ADE@8sec and FDE@8sec of above listed implementations. It is observed that the proposed three-channel framework (HEAT-I-R) outperforms its  two channel (HEAT-R) and one-channel (HEAT) ablations, which supports the intuition that individual dynamics, inter-agent interactions, and road information all benefit trajectory prediction. It is also noticed that one-channel GAT based method (GAT in Tab.~\ref{tab: itablation}) performs as poor as the non-interaction-aware method (R in Tab.~\ref{tab: itablation}.), but their combination (GAT-R) improves prediction accuracy. GAT-based methods are not suitable for trajectory prediction with exclusive coordinate systems, where the spatial relationships of agents are stored in edge features, because GATs ignore edge features.

Fig.~\ref{fig: boxplotFT} shows box plots of the TDE of three ablatives models (GAT, HEAT, and HEAT-I-R) over a eight-second prediction horizon. The red boxes shows the results of GAT, the green boxes the results of HEAT, and blue boxes the result of the proposed HEAT-I-R. Outliers and the results of other ablative models is not plotted for clarity. It can be seen that the proposed HEAT-based model shows more stable performance (with shorter interquartile range (IQR)) than the GAT-based model, and the proposed three channel framework (HEAT-I-R) further improves accuracy and stability.

\subsection{Validation On Homogeneous Dataset}
The proposed HEAT is an immediate extension to GATs~\cite{velickovic2018graph} while the GAT is designed for homogeneous graph. To compare the proposed HEAT with GAT on the single trajectory prediction task, this section constructs two homogeneous datasets using vehicle trajectories provided by the public accessible NGSIM US-101~\cite{ushighway101} dataset.
\subsubsection{Homogeneous Dataset}
 The trajectories in NGSIM US-101 dataset are recorded from a  segmentation of U.S. Highway 101 in 10 Hz. Since most of the trajectories did not change lane throughout the study area, this work reprocess the dataset to select roughly balanced dataset where lane-keeping trajectories do not dominate the dataset. The dataset is constructed by first selecting target vehicles which have changed its lane only once during recording, then select trajectory segments for the target vehicle and its neighboring vehicles. The data processing procedure is similar to~\cite{mo2020interaction} except that this work allows an arbitrary number of neighboring vehicles. After above processing, totally $63,176$ data pieces are selected for training ($53,176$) and validation ($10,000$). The selected data is further processed to formulate two different datasets, the one with exclusive coordinate system, and the other one with shared coordinate system. In the former dataset, each agent is placed in its own own coordinate system, in which the agent's current position and yaw angle are zeros, while in the later one, all the agents share the coordinate system whose origin is fixed at the current position of the target vehicle and horizontal axis points to the direction of the target vehicle's current velocity.
The former dataset has edge attributes containing the relative position of the agent in the source node to the agent in the target node, while the later one does not have edge attributes because of the shared stationary frame of reference.

\subsubsection{Ablative Models}
The following models are implemented to show the effectiveness the proposed HEAT layer:
\begin{itemize}
    \item R: One-channel model considering the target vehicle's RNN-encoded  dynamics feature only for future trajectory prediction.
    \item GAT: One-channel model predicting the future trajectory of the target vehicle considering its graph-modeled interaction feature extracted using GAT.
    \item GAT-R: Two-channel model jointly considering the GAT-extracted interaction and RNN-encoded dynamics features for trajectory prediction.
    \item HEAT: One-channel model predicting the future trajectory of the target vehicle considering its graph-modeled interaction feature extracted using HEAT.
    \item HEAT-R: Two-channel model jointly considering the HEAT-extracted interaction and RNN-encoded dynamics features for trajectory prediction.
\end{itemize}
All the above models are trained and validated on the dataset with exclusive coordinate system and the GAT-based baselines are further implemented in the dataset with shared coordinate system.

\subsubsection{Compared Models}
The proposed model is compare with the following baselines in the literature: 
\begin{itemize}
    \item GAIL-GRU: A generative adversarial imitation learning model described in~\cite{kuefler2017imitating}, which has access to ground-truth trajectories of it surrounding vehicles provided by the NGSIM dataset when predicting.
    \item CS-LSTM: An interaction-aware trajectory prediction model proposed in~\cite{deo2018convolutional}, which represents the inter-vehicular interaction as an occupancy grid, whose cells containing the corresponding vehicles' dynamics features extracted using LSTMs. A aggregation method called convolutional social pooling is proposed to extract interaction features from the occupancy grid.
    \item CS-LSTM(M): An extension to CS-LSTM proposed in ~\cite{deo2018convolutional}, which generates multi-modal predictions conditional on six pre-defined maneuvers. This method has access to the maneuver labels. 
    \item GRIP: A graph-based method proposed in~\cite{li2019grip}, which represents interaction of near vehicles as a graph and designs several graph convolutional blocks to extract the interaction features. 
    \item MATF-GAN: Multi-Agent Tensor Fusion proposed in~\cite{zhao2019multi}, which applies convolutional fusion to extract multi-agent interactions and retains the scene context by spatially aligning interactions tensor with the context feature map.
    \item CNN-LSTM: An interaction-aware trajectory prediction model proposed in~\cite{mo2020interaction}, which considers the target vehicle's interaction with its eight closes neighbors using a 3$\times$3 grid, rather than consider all the vehicles falling in a predefined neighborhood.
    \item Scale-Net: A graph-based multi-agent trajectory predictor proposed in~\cite{jeon2020scale}, which models the inter-vehicular interaction as an edge-featured homogeneous graph and applies edge-featured graph neural networks~\cite{gong2019exploiting} for feature extraction. 
\end{itemize}

\subsubsection{Metric}
The prediction results in this section are reported in terms of root-mean-square error (RMSE) in meters to compare with existing works~\cite{mo2020interaction, deo2018convolutional, jeon2020scale}.
\begin{equation}
    RMSE(t_p) = \sqrt{\frac{1}{n}\sum^{n}_{i=1} \left((\hat{x}^i_{t_p} - x^i_{t_p} )^2 + (\hat{y}^i_{t_p} - y^i_{t_p} )^2 \right)},
    \label{eq: metric}
\end{equation}
where $n=10000$ is the number of data for validation, $(\hat{x}^i_{t_p}, \hat{y}^i_{t_p})$ is the predicted position of the target vehicle in data $i$ at time ${t_p}$, and $(x^i_{t_p}, y^i_{t_p})$ is the corresponding ground truth.

\subsubsection{Implementation Details}
All the models listed in Tab.~\ref{tab: ngablation} are implemented using PyTorch~\cite{NEURIPS2019_9015} and the GAT-based methods are implemented with PyTorch Geometric~\cite{Fey/Lenssen/2019} in addition to PyTorch.

\subsubsection{Results} 
Tab.~\ref{tab: othersresults} compares the proposed method with existing works. It shows that the proposed HEAT-R method outperforms existing models at longer prediction horizons (3-5 sec) and matches the state-of-the-art methods in short-term prediction (1-2 sec). It can be seen that the accuracy of the proposed HEAT-R is quite close to that of CNN-LSTM~\cite{mo2020interaction} which uses a share coordinate system and accepts exactly eight close neighboring vehicle, while HEAT-R handles an arbitrary number of neighboring vehicles and uses exclusive coordinate system, which standardises the input sequence and narrows down the search space for the input encoder.

Tab.~\ref{tab: ngablation} compares HEAT with GAT-based and RNN-based methods, where the first five rows (\#1-5) show the results on the dataset with exclusive coordinate system, and the last two rows (\#6-7) show the results of GAT-based models on the dataset with shared coordinate system. It can be seen that the graph-based interaction-aware methods (\#2-7) outperforms the non-interaction-aware RNN-based one (\#1). This observation is consistent with previous works~\cite{deo2018convolutional, zhao2019multi, mo2020interaction}, which shows again the necessity to model interaction for trajectory prediction. The GAT-based methods show better performance on the second dataset (shared, \#6,7) comparing to the results on the first dataset (exclusive, \#2,3). This is quite reasonable in that GAT ignores the spatial relationship among agents contained in the edge features in the first dataset, while the spatial relationship is preserved by the shared coordinate system in the second dataset. The proposed HEAT-based methods (\#4-5) for the dataset with exclusive coordinate system outperform all GAT-based methods. This shows the advantage of using exclusive coordinate system and applying HEAT for interaction extraction.

\begin{table}[!ht]
\caption{\textbf{Prediction performance comparison with existing works (RMSE in meters)}}
\centering
\begin{tabular}{|c|c|c c c c c|} 
\hline
&\multirow{2}{6em}{ \textbf{Methods }} &\multicolumn{5}{c|}{\textbf{Prediction horizon}}\\
\cline{3-7} 
& & \textbf{\textit{1 sec}}& \textbf{\textit{2 sec}}& \textbf{\textit{3 sec}} & \textbf{\textit{4 sec}} & \textbf{\textit{5 sec}} \\
\hline
1 & HEAT-R (Ours) & 0.68 & 0.92 & \textbf{1.15} & \textbf{1.45}  & \textbf{2.05} \\
\hline
2 & CS-LSTM~\cite{deo2018convolutional} & 0.61 & 1.27 & 2.09 & 3.10 & 4.37 \\
\hline
3 & GRIP~\cite{li2019grip} & \textbf{0.37} & \textbf{0.86} & 1.45 & 2.21 & 3.16 \\
\hline
4 & CNN-LSTM~\cite{mo2020interaction} & 0.64 & 0.96 & 1.22 & 1.53 & 2.09 \\
\hline
5 & CS-LSTM(M)~\cite{deo2018convolutional} & 0.62 & 1.29 & 2.13 & 3.20 & 4.52 \\
\hline
6 & GAIL-GRU~\cite{kuefler2017imitating} & 0.69 & 1.51 & 2.55 & 3.65 & 4.71 \\
\hline
7 & Scale-Net~\cite{jeon2020scale} & 0.46 & 1.16 & 1.97 & 2.91 & - \\
\hline
8 & MATF-GAN~\cite{zhao2019multi} & 0.66 & 1.34 & 2.08 & 2.97 & 4.13 \\
\hline
\end{tabular}
\vspace{0.2cm}
\label{tab: othersresults}
\end{table}

\begin{table}[!ht]
\caption{\textbf{Prediction performance comparison over ablative implementations (RMSE in meters)}}
\centering
\begin{tabular}{|c|c|c c c c c|} 
\hline
&\multirow{2}{4em}{ \textbf{Methods }} &\multicolumn{5}{c|}{\textbf{Prediction horizon}}\\
\cline{3-7} 
& & \textbf{\textit{1 sec}}& \textbf{\textit{2 sec}}& \textbf{\textit{3 sec}} & \textbf{\textit{4 sec}} & \textbf{\textit{5 sec}} \\
\hline
1 & R-1 & 0.6931 & 1.7275 & 3.0850 & 4.7735 & 6.7855 \\
\hline
2 & GAT-1 & 0.7808 & 1.4916 & 2.3914 & 3.4981 & 4.8713 \\
\hline
3 & GAT-R-1 & 0.8228 & 1.6987 & 2.7385 & 3.9672 & 5.4129 \\
\hline
4 & HEAT-1 & 0.6590 & 0.8556 & 1.0469 & 1.3216 & 1.8894 \\
\hline
5 & HEAT-R-1 & 0.6794 & 0.9212 & 1.1528 & 1.4457,  & 2.0518 \\
\hline
\hline
6 & GAT-2  & 0.7685 & 1.2115 & 1.7343 & 2.4533 & 3.5466 \\
\hline
7 & GAT-R-2 & 0.6439 & 0.9592 & 1.2643 & 1.6489 & 2.3124 \\
\hline
\end{tabular}
\vspace{0.2cm}
\label{tab: ngablation}
\end{table}

\section{Conclusion}
\label{sec: conclusion}
This work proposes a three-channel framework for simultaneous heterogeneous multi-agent trajectory prediction. It represents inter-agent interaction in traffic with a directed edge-featured heterogeneous graph, designs a novel heterogeneous edge-enhanced graph attention network for inter-agent interaction modeling, introduces gate mechanism for selective map sharing across all target agents. Validations on both the urban driving dataset (INTERACTION) and the highway driving dataset (NGSIM) show that the proposed method achieves state-of-the-art performance while simultaneous predict multi-agent trajectories of an arbitrary number of heterogeneous agents. 

One immediate improvement to the proposed method is to handle the multi-modality of traffic participants' behaviors by introducing multi-modal prediction and handling multi-modality will largely reduce the minimum ADE. Another way is to incorporate rich infrastructure information, such as traffic lights for trajectory prediction.

\section*{Appendix}
\begin{figure*}
    \centering
    \includegraphics[trim={0cm 0cm 0cm 0cm}, clip, width=1.0\textwidth]{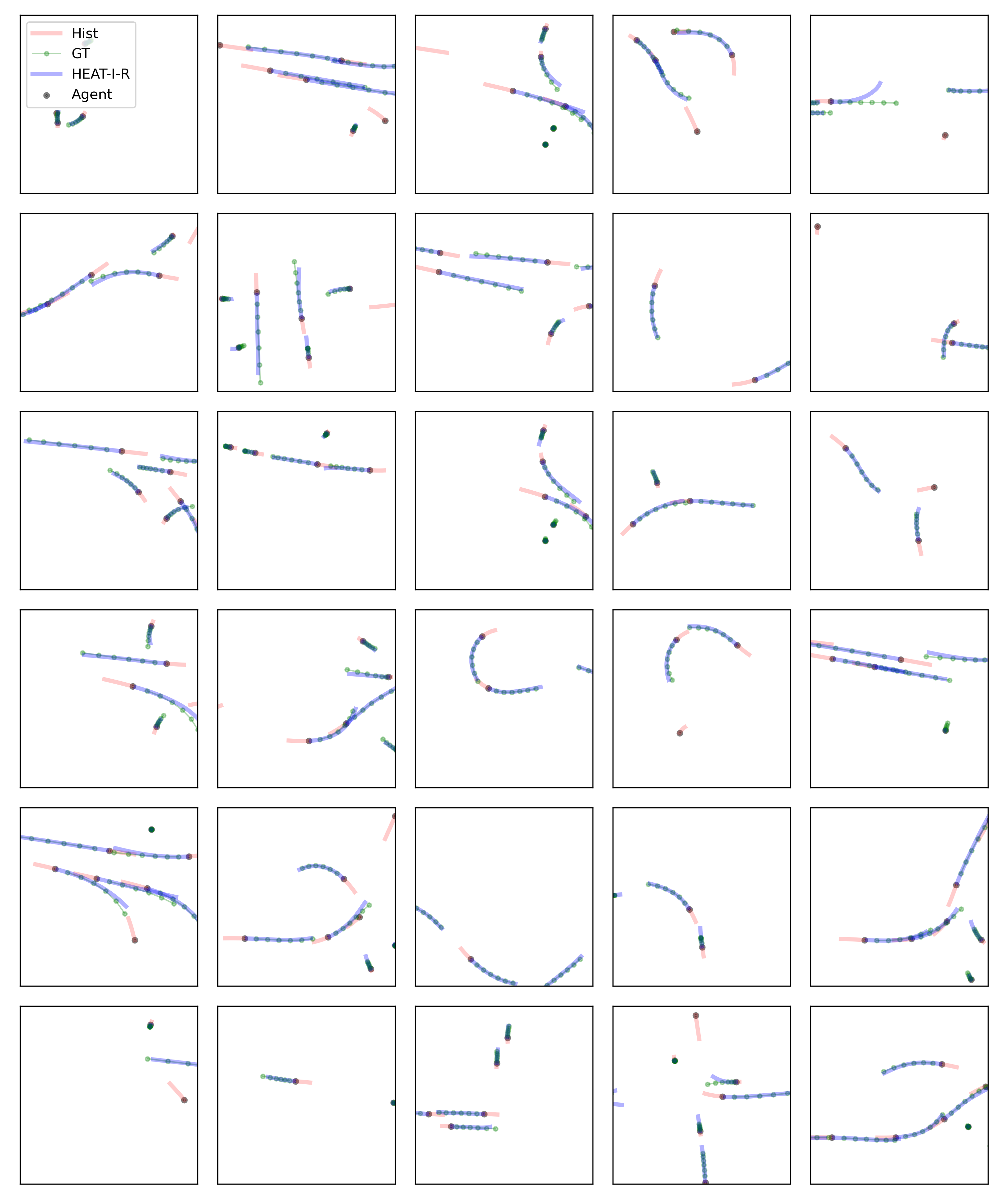}
    \vspace{-8mm}
    \caption{\textbf{Visualized prediction results.} This figure visualizes prediction results of the proposed HEAT-based multi-agent trajectory prediction method on various driving scenarios randomly sampled from the entire INTERACTION dataset.
    Black dots are the agents' (could be either vehicle or pedestrian/bicyclist) current position, red lines are their one-second historical trajectories, green lines with dots are their ground truth future trajectories over three seconds, and blue lines are the trajectories predicted by the proposed framework with HEAT.
    It shows that the proposed method is able to simultaneously predict trajectories of variable number of heterogeneous agents in different scenarios.}
    \label{fig: visualization}
\end{figure*}

Prediction results of the proposed framework with HEAT on thirty scenarios randomly sampled from the entire INTERACTION dataset are shown in Fig.~\ref{fig: visualization}.

\section*{Acknowledgment}
The authors would like to thank...

\ifCLASSOPTIONcaptionsoff
  \newpage
\fi

\bibliographystyle{IEEEtran}
\bibliography{reference.bib}

\end{document}